\documentclass[10pt,twocolumn,letterpaper]{article}

\usepackage{iccv}
\usepackage{times}
\usepackage{epsfig}
\usepackage{graphicx}
\usepackage{amsmath}
\usepackage{amssymb}
\usepackage{dsfont}
\usepackage{color}
\usepackage{comment}
\usepackage{subfig}
\usepackage{wrapfig}
\usepackage{nicefrac}
\usepackage{float}
\usepackage[normalem]{ulem}
\usepackage{animate}
\usepackage[export]{adjustbox}

\usepackage[pagebackref=true,breaklinks=true,letterpaper=true,colorlinks,bookmarks=false]{hyperref}

\iccvfinalcopy 

\def\iccvPaperID{238} 

\ificcvfinal\fi

\definecolor{darkgreen}{RGB}{0,127,0}
\definecolor{darkblue}{RGB}{0,0,127}
\definecolor{darkred}{RGB}{127,0,0}
\definecolor{darkmagenta}{RGB}{127,0,127}
\definecolor{darkcyan}{RGB}{0,127,127}
\definecolor{darkyellow}{cmyk}{0,0.7,0.9,0.3}

\newcommand{\I}{\mathcal{I}}
\newcommand{\NV}{\mathcal{I}^{\text{NV}}}
\newcommand{\NVI}{\widetilde{\mathcal{I}}^{\text{NV}}}

\newcommand{\D}{\mathcal{D}}
\newcommand{\DNV}{\mathcal{D}^{\text{NV}}}
\newcommand{\R}{\mathcal{R}}

\begin{document}

\title{Extreme View Synthesis}

\author{Inchang Choi$^{1,2}$
\and
Orazio Gallo$^{1}$
\and
Alejandro Troccoli$^{1}$
\and
Min H. Kim$^{2}$
\and
Jan Kautz$^{1}$
\and
$^{1}$NVIDIA
\and
$^{2}$KAIST
}
\twocolumn[{%
\vspace{-7mm}
\renewcommand\twocolumn[1][]{#1}%
\maketitle
\begin{center}
    \centering
    \captionsetup{type=figure}
    \includegraphics[width=\textwidth]{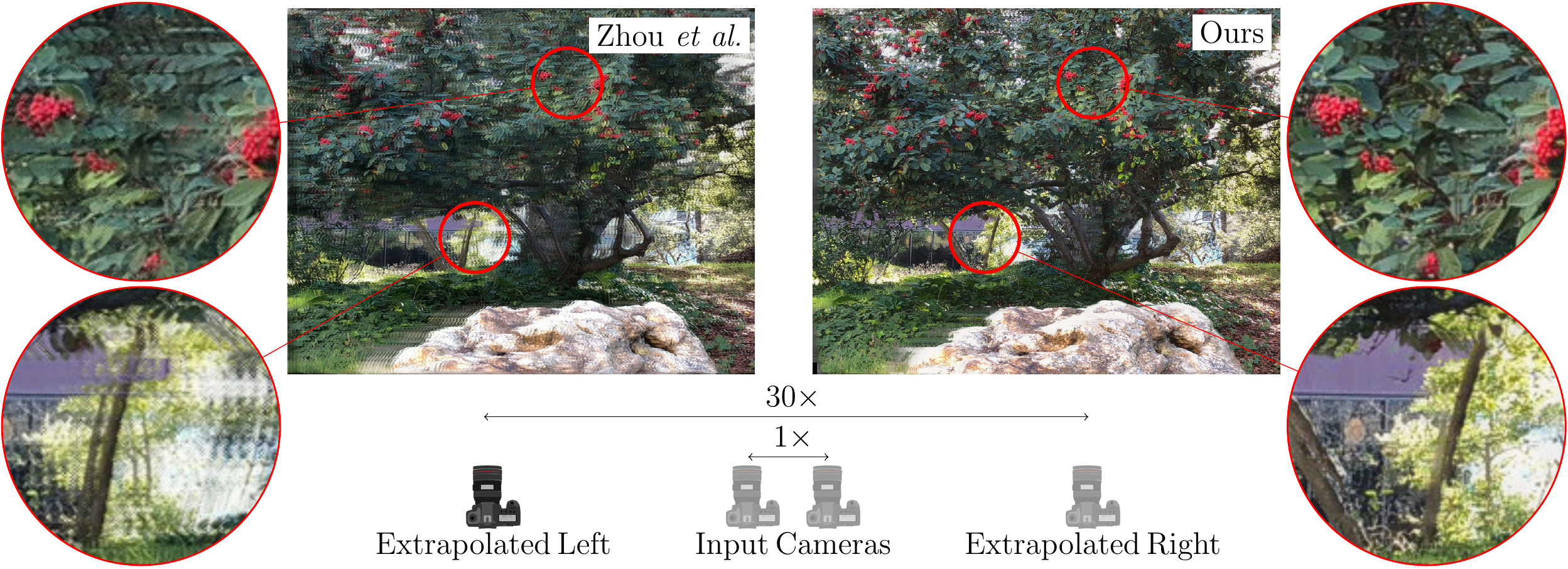}
    \small{\captionof{figure}{We propose a novel view synthesis method that can generate \emph{extreme} views, \ie, images synthesized from a small number of cameras (two in this example) and from significantly different viewpoints. In this comparison with the method by Zhou~\etal~\cite{zhou2018stereo}, we show the left view from the camera setup depicted above. Even at a $30\times$ baseline magnification our method produces sharper results.}\label{fig:teaser}}
\end{center}%
}]


\begin{abstract}
We present Extreme View Synthesis, a solution for novel view extrapolation that works even when the number of input images is small---as few as two.
In this context, occlusions and depth uncertainty are two of the most pressing issues, and worsen as the degree of extrapolation increases.
We follow the traditional paradigm of performing depth-based warping and refinement, with a few key improvements. 
First, we estimate a depth probability volume, rather than just a single depth value for each pixel of the novel view.
This allows us to leverage depth uncertainty in challenging regions, such as depth discontinuities.
After using it to get an initial estimate of the novel view, we explicitly combine learned image priors and the depth uncertainty to synthesize a refined image with less artifacts.
Our method is the first to show visually pleasing results for baseline magnifications of up to $30\times$.
The code is available at \url{https://github.com/NVlabs/extreme-view-synth}
\end{abstract}

\vspace{-4mm}
\section{Introduction}\label{sec:intro}

The ability to capture visual content and render it from a different perspective, usually referred to as \emph{novel view synthesis}, is a long-standing problem in computer graphics.
When appropriately solved, it enables telepresence applications such as head-mounted virtual and mixed reality, and navigation of remote environments on a 2D screen---an  experience popularized by Google Street View.
The increasing amount of content that is uploaded daily to sharing services offers a rich source of data for novel view synthesis.
Nevertheless, a seamless navigation of the virtual world requires a more dense sampling than these sparse observations offer.
Synthesis from sparse views is challenging, in particular when generating views creating disocclusions, a common situation when the viewpoint is extrapolated, rather than interpolated, from the input cameras.

\begin{figure}[t!]
    \vspace{-3mm}
    \centering
    \input{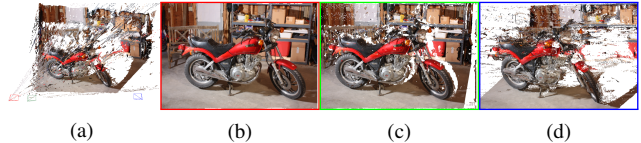}
    \caption{(a) A point cloud and three cameras. (b)-(d) The images ``captured'' from the red, green, and the blue cameras. The point cloud was generated from the depth map of the red camera. Depth uncertainty causes larger artifacts as the viewpoint moves farther from the red camera.}
    \label{fig:depthUncertainty}
\end{figure}

Early novel view synthesis methods can generate new images by interpolation either in pixel space~\cite{chen1993view}, or in ray space~\cite{levoy1996light}.
Novel views can also be synthesized with methods that use 3D information explicitly.
A typical approach would use it to warp the input views to the virtual camera and merge them based on a measure of quality~\cite{buehler2001unstructured}.
The advantage of such methods is that they explicitly leverage geometric constraints.
Depth, however, does not come without disadvantages. First and foremost is the problem of occlusions. Second, depth estimation is always subject to a degree of uncertainty.
Both of these issues are further exacerbated when the novel view is pushed farther from the input camera, as shown in Figure~\ref{fig:depthUncertainty}.
Existing methods deal with uncertainty by propagating reliable depth values to similar pixels~\cite{CDSD13}, or by modeling it explicitly~\cite{penner2017soft}.
But these approaches cannot leverage depth to refine the synthesized images, nor do they use image priors to deal with the unavoidable issues of occlusions and artifacts.

More recent approaches use large data collections and learn the new views directly~\cite{flynn2016deepstereo,zhou2018stereo}.
The power of learning-based approaches lies in their ability to leverage image priors to fill missing regions, or correct for poorly reconstructed ones.
However, they still cause artifacts when the position of the virtual camera differs significantly from that of the inputs, in particular when the inputs are few.

In their \emph{Stereo Magnification} work, Zhou~\etal cleverly extract a layered representation of the scene~\cite{zhou2018stereo}.
The layers, which they learn to combine into the novel view, offer a regularization that allows for an impressive stereo baseline extrapolation of up to $4.5\times$.
Our goal is similar, in that we want to use as few as two input cameras and extrapolate a novel view.
Moreover, we want to push the baseline extrapolation much further, up $30\times$, as shown in Figure~\ref{fig:teaser}.
In addition, we allow the virtual camera to move and rotate freely, instead of limiting to translations along the baseline.

At a high level, we follow the depth-warp-refine paradigm, but we leverage two key insights to achieve such large extrapolation.
First, depth estimation is not always reliable: instead of exact depth estimates, we use depth probability volumes.
Second, while image refinement networks are great at learning generic image priors, we also use explicit information about the scene by sampling patches according to the depth probability volumes. 
By combining these two concepts, our method works for both view interpolation and extreme extrapolation.
We show results on a large number of examples in which the virtual camera significantly departs from the original views, even when only two input images are given.
To the best of our knowledge, ours is the first method to produce visually pleasing results for such extreme view synthesis from unstructured cameras.

\section{Related Work}\label{sec:related}

\begin{figure*}
  \centering
    \includegraphics[width=.8\textwidth,trim={0 5 0 7},clip]{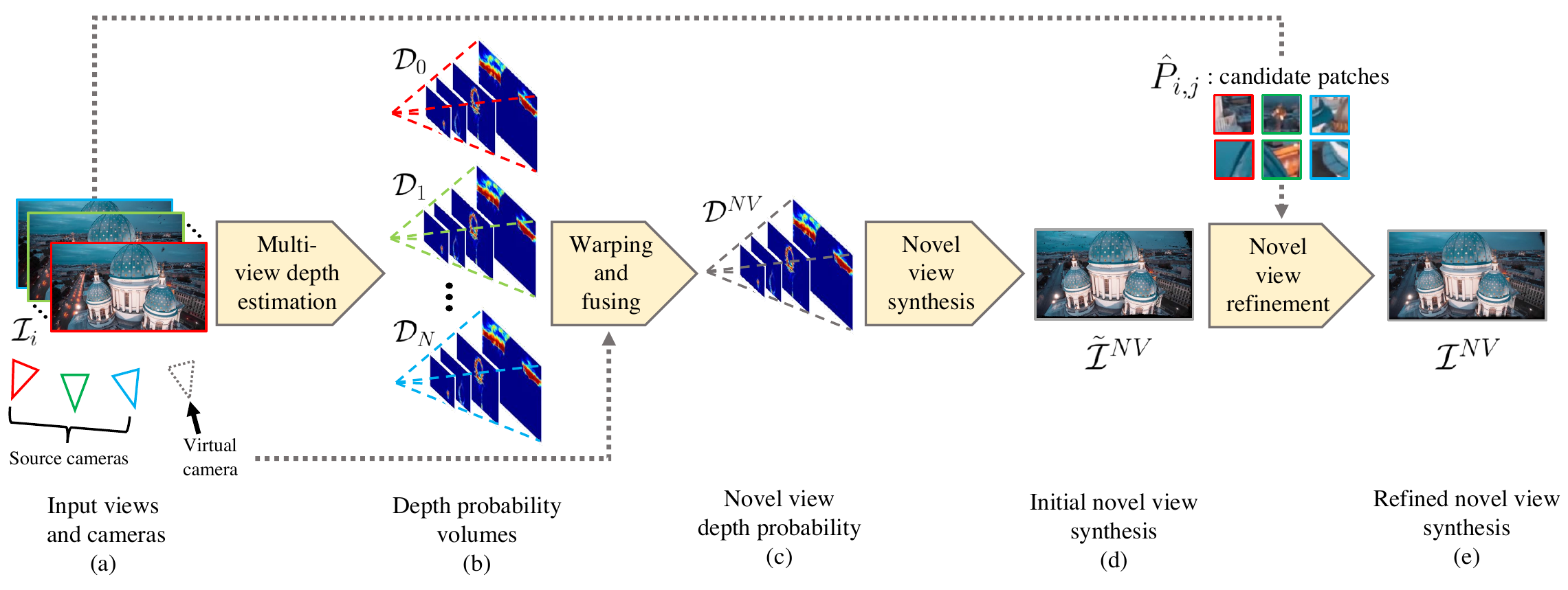}%

    \caption[]{\label{fig:overview}%
    Method overview: from a set of posed input views (a), we generate a set of depth probability volumes for each view (b). Given the novel view camera pose, we create its depth probability volume via warping and fusion of the input depth volumes (c). Next, we synthesize an initial novel view (d), which we refine with a neural network to synthesize the final image (e). Our image refinement is done in a patch-based manner guided by the depth distribution.}
    \vspace{-3mm}
\end{figure*}

Early methods for novel view synthesis date back several decades~\cite{greene1986environment}.
Image interpolation methods, among the first approaches to appear, work by interpolating between corresponding pixels from the input images~\cite{chen1993view}, or between rays in space~\cite{levoy1996light}.
The novel view can also be synthesized as a weighted combination of the input cameras, when information about the scene geometry is available~\cite{buehler2001unstructured,debevec1996modeling}. All of these methods generally assume additional information---correspondences, depth, or geometry---to be given.

Recent methods produce excellent results taking only images as an input.
This can be done, for instance, by using an appropriate representation of the scene, such as plane sweep volumes, and by learning weights to merge them down into a single image~\cite{flynn2016deepstereo}. Further building on the concept layered depth images~\cite{he1998layered}, Zitnick~\etal developed a
high-quality video-based rendering system for dynamic scenes that can interpolate between views~\cite{Zitnick2004HighqualityVV}.
Zhou~\etal propose a learned layer-based representation of the scene, dubbed MPI~\cite{zhou2018stereo}.
Their results are impressive, but quickly degrade beyond limited translations of the novel view.
The works of Mildenhall~\etal~\cite{mildenhall2019local} and Srinivasan~\etal~\cite{srinivasan2019pushing} build on the MPI representation further improving the quality of the synthesized view, even for larger camera translations\footnote{These works were published after the submission of this paper and are included here for a more complete coverage of the state-of-the-art.}.

A different approach is to explicitly use depth information, which can be estimated from the input images directly, and used to warp the input images into the novel view.
Kalantari~\etal, for instance, learn to estimate both disparity and the novel view from the sub-aperture images of a lightfield camera~\cite{kalantari2016learning}.
For larger displacements of the virtual camera, however, depth uncertainty results in noticeable artifacts.
Chaurasia~\etal take accurate but sparse depth and propagate it using super-pixels based on their similarity in image space~\cite{CDSD13}.
Penner and Zhang explicitly model the confidence that a voxel corresponds to empty space or to a physical surface, and use it while performing back-to-front synthesis of the novel view~\cite{penner2017soft}.

The ability of deep learning techniques to learn priors has also paved the way to single-image methods.
Srinivasan~\etal learn a light field and depth along each ray from a single image~\cite{srinivasan2017learning}.
Zhou~\etal cast this problem as a prediction of appearance flows, which allows them to synthesize novel views of a 3D object or scene from a single observation~\cite{zhou2016view}.
From a single image, Xie~\etal produce stereoscopic images~\cite{xie2016deep3d}, while Tulsiani~\etal infer a layered representation of the scene~\cite{tulsiani2018layer}.

Our approach differs from published works for its ability to generate extrapolated images under large viewpoint changes and from as few as two cameras.

\section{Overview}\label{sec:overview}

Our goal is to synthesizes a novel view, $\NV$, from $N$ input views, $\I_i$.
A common solution to this problem is to estimate depth and use it to warp and fuse the inputs into the novel view.
However, depth estimation algorithms struggle in difficult situations, such as regions around depth discontinuities; this causes warping errors and, in turn, artifacts in the final image.
These issues further worsen when $N$ is small, or $\NV$ is extrapolated, \ie, when the virtual camera is not on the line connecting the centers of any two input cameras.
Rather than using a single depth estimate for a given pixel, our method accounts for the depth's probability distribution, which is similar in spirit to the work of Liu~\etal~\cite{liu2019neural}.
We first estimate $N$ distributions $\D_i$, one per input view, and combine them to estimate the distribution for the virtual camera, $\DNV$, Section~\ref{sec:depth}. Based on the combined distribution $\DNV$, we render the novel view back to front, Section~\ref{sec:single_synth}.
Finally, we refine $\NV$ at the patch level informed by relevant patches from the input views, which we select based on the depth distribution and its uncertainty, Section~\ref{sec:refinement}. Figure~\ref{fig:overview} shows an overview of the method.

\section{Estimating the Depth Probability Volume}\label{sec:depth}
Several methods exist that estimate depth from multiple images~\cite{Kar2017LearningAM,galliani2015massively}, stereo pairs~\cite{kendall17deepstereo, khamis18stereonet}, and even single image~\cite{MegaDepthLi18,saxena2006learning}.
Inspired by the work of Huang~\etal, we treat depth estimation as a learning-based, multi-class classification problem~\cite{huang2018deepmvs}.
Specifically, depth can be discretized into $n_d$ values and each depth value can be treated as a class.
Depth estimation, then, becomes a classification problem: each pixel $(x_i,y_i)$ in $\I_i$ can be associated with a probability distribution over the $n_d$ depth values along $\R_i(x_i,y_i)$, the ray leaving the camera at $(x_i,y_i)$ and traversing the scene.
We refer to the collection of all the rays for camera $i$ as a \emph{depth probability volume}, $\D_i \in \mathbb{R}^{h\times w \times n_d}$, where $h\times w$ is the resolution of $\I_i$.
The network to estimate the $\D_i$'s, can be trained with a cross-entropy loss against ground truth one-hot vectors that are 1 for the correct class and 0 elsewhere, as in Huang~\etal~\cite{huang2018deepmvs}.
We follow the common practice of uniformly sampling disparity instead of depth\footnote{Technically, ``disparity'' is only defined in the case of a stereo pair. Here we use the term loosely to indicate a variable that is inversely proportional to depth.} to improve the estimation accuracy of closer objects.

\begin{figure}[t!]
    \centering
    \includegraphics[width=\columnwidth,trim={0 140mm 200mm 0}, clip]{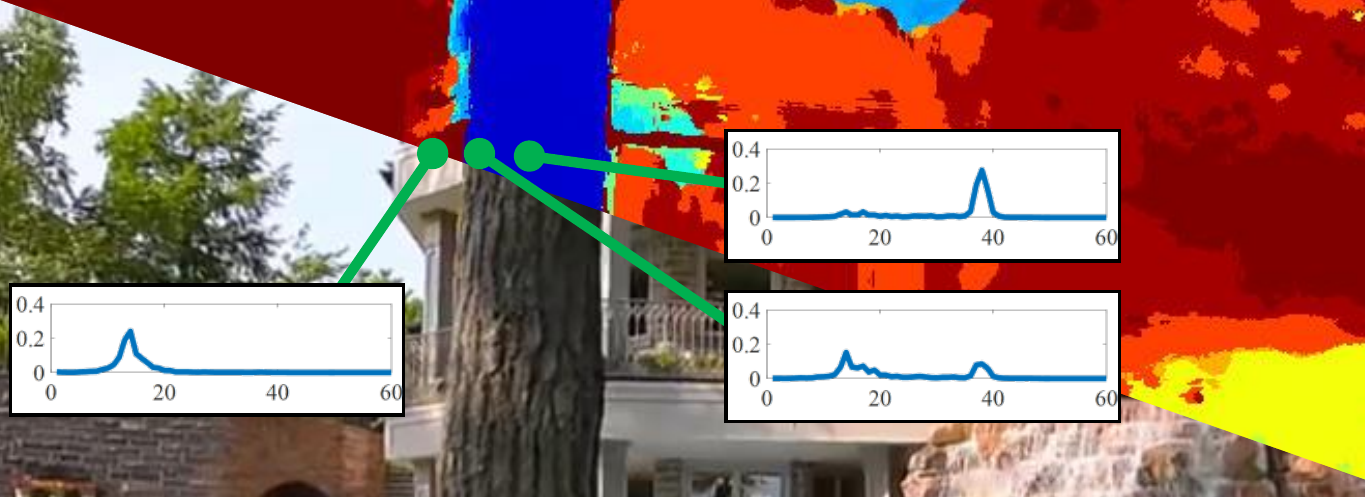}
    \caption{Depth probability distributions along three rays in $\mathcal{D}$. The disparity shows clear peaks for points that are  sufficiently distant from a depth discontinuity. Closer to the edge, the inherent uncertainty is captured by the presence of two lower peaks: one corresponding to the foreground, and one to the background.}
    \label{fig:depth_prob}
\end{figure}

Empirically, we observe that the resulting depth volumes exhibit desirable behaviors.
For most regions, the method is fairly certain about disparity and the probability along $\R_i(x,y)$ presents a single, strong peak around the correct value.
Around depth discontinuities, where the point-spread-function of the lens causes pixels to effectively belong to both foreground and background, the method tends to produce a multi-modal distribution, with each peak corresponding to the disparity levels of the background and foreground, see for instance Figure~\ref{fig:depth_prob}.
This is particularly important because depth discontinuities are the most challenging regions when it comes to view synthesis.

Solving for the depth probability volumes requires that we know the location and the camera's intrinsic parameters for each input view.
We estimate these using Colmap~\cite{colmap}.
For a given scene, we set the closest and farthest disparity levels as the bottom $2$ and top $98$ depth percentiles respectively, and use $n_d=100$ uniformly spaced disparity steps.
Similarly to the method of Huang~\etal, we also cross-bilateral filter the depth probability volume guided by an input RGB image~\cite{kraehenbuehl2011crf}. 
However, we find $\theta_{\alpha}=25$, $\theta_{\beta}=10$, and $\mu=5$ to work better for our case and iterate the filter for $5$ times.
We refer the reader to Kr{\"a}henh{\"u}hl and Koltun for the role of each parameter~\cite{kraehenbuehl2011crf}.

Finally, we can estimate the probability volume $\DNV$ for the novel view by resampling these probability volumes.
Conceptually, the probability of each disparity $d$ for each pixel $(x,y)$, $\DNV(x,y,d)$, can be estimated by finding the intersecting rays $\R_i$'s from the input cameras and average their probability.
This, however, is computationally demanding.
We note that this can be done efficiently by resampling the $\D_i$'s with respect to $\DNV$, accumulating each of the $\D_i$  volumes into the novel view volume, and normalizing by the number of contributing views.
This accumulation is sensible because the probability along $\R_i$ is a proper distribution.
This is in contrast with traditional cost volumes~\cite{hosni2012fast} for which costs are not comparable across views:
the same value for the cost in two different views may not indicate that the corresponding disparities are equally likely to be correct.
Depth probability volumes also resemble the soft visibility volumes by Penner and Zhang~\cite{penner2017soft}.
However, their representation is geared towards identifying empty space in front of the first surface.
Therefore, they behave differently in regions of uncertainty, such as depth discontinuities, where depth probability volumes carry information even beyond the closest surface.

\begin{figure}
    \centering
    \includegraphics[width=\columnwidth, trim={0 4 0 8}, clip]{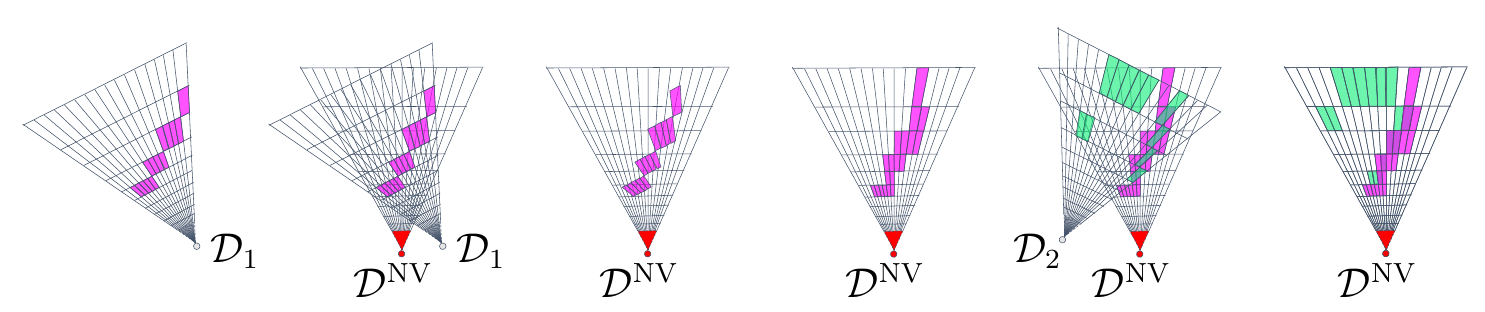}
    \caption{To compute the depth probability volume with respect to the novel view, we resample the input volumes and  accumulate them. Here we only look at a planar slice of the depth probability volumes, and we make the simplifying assumption that the input volumes have $p=1$ for one disparity and $p=0$ for all the others. Note that the probability along the rays in the final result do not sum to 1 and, therefore require an additional normalization.}
    \label{fig:depth_projection}
    \vspace{-5mm}
\end{figure}

Figure~\ref{fig:depth_projection} shows an example of the resampling procedure, where we consider only a planar slice of the volumes and, for simplicity, that the probability along the input rays is binary.
We use nearest neighbor sampling, which, based on our experiments, yields quality comparable with tri-linear interpolation at a fraction of the cost.
After merging all views, we normalize the values along each ray in $\DNV$ to enforce a probability distribution.

\section{Synthesis of a Novel View}\label{sec:single_synth}

Using the depth probability volume $\DNV$, we backward warp pixels from the inputs $\I_i$ and render in a back-to-front fashion an initial estimate of the novel view, $\NVI$.
Specifically, we start from the farthest plane, where $d=0$, and compute a pixel in the novel view as 
\begin{equation}
    \left.\NVI(x,y)\right|_{d=0} = \text{R}\left(\left\{\I_i(x_i,y_i) \cdot \mathds{1}_{\{\DNV(x,y,0) > t\}}\right\}_{i=1:N}\right),\label{eq:d0render}
\end{equation}
where $\mathds{1}$ is the indicator function, and $(x_i,y_i)$ are the coordinates in $\I_i$ that correspond to $(x,y)$ in $\NVI$. Note that these are completely defined by the cameras' centers and the plane at $d$.
$\text{R}$ is a function that merges pixels from $\I_i$ weighting them based on the distance of the cameras' centers, and the angles between the cameras' principal axes. Details about the threshold $t$ and the weights are in the Supplementary.
As we sweep the depth towards a larger disparity $d$, \ie, closer to the camera, we overwrite those pixels for which $\DNV(x,y,d)$ is above threshold\footnote{An alternative to overwriting the pixels, is to weigh their RGB values with the corresponding depth probabilities.
However, in our experiments, this resulted in softer edges or ghosting that were harder to fix for the refinement network (Section~\ref{sec:refinement_network}).
We speculate that the reason to be that such artifacts are more ``plausible'' to the refinement network than abrupt and incoherent RGB changes.}.

\begin{figure*}[tb]
    \centering
    \includegraphics[width=\textwidth]{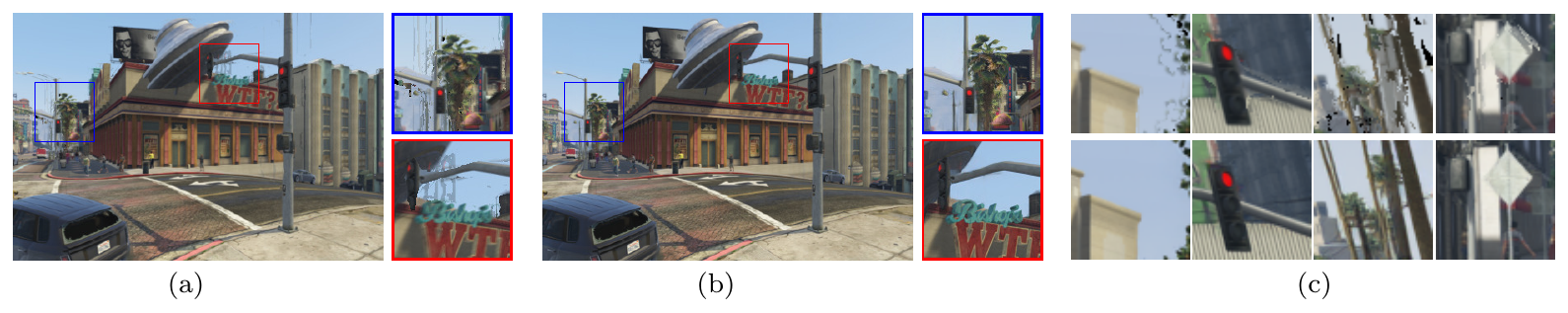}
    \vspace{-8mm}
    \caption{The novel view $\widetilde{\mathcal{I}}^{\text NV}$, obtained by just warping the inputs, presents several types of artifacts (a). Our refinement network uses the depth probability as well as patches from the input images to fix them (b). More examples of synthesized (top) and refined (bottom) patches are shown in (c).}
    \label{fig:stages}
    \vspace{-3mm}
\end{figure*}
The resulting image $\NVI$ will, in general, presents artifacts and holes, see Figure~\ref{fig:stages}(a).
This is expected, since we are rejecting depth estimates that are too uncertain, and we overwrite pixels as we sweep the depth plane from back to front.
However, at this stage we are only concerned with generating an initial estimate of the novel view that obeys the geometric constraints captured by the depth probability volumes.

\section{Image Refinement}\label{sec:refinement}
The image $\NVI$ synthesized as described in Section~\ref{sec:single_synth} is generally affected by apparent artifacts, as shown in Figures~\ref{fig:stages}(a) and (c).
Most notably, these include regions that are not rendered, either because of occlusions or missing depth information, and the typical ``fattening'' of edges at depth discontinuities.
Moreover, since we render each pixel independently, structures may be locally deformed.
We address these artifacts by training a refinement network that works at the patch level.
For a pixel $p$ in $\NVI$, we first extract a $64\times 64$ patch $\widetilde{P}^{\text{NV}}$ around it (for clarity of notation, we omit its dependence on $p$).
The goal of the refinement network is to produced a higher quality patch with less artifacts.
One could consider the refinement operation akin to denoising, and train a network to take a patch $\widetilde{P}^{\text{NV}}$ and output the refined patch, using a dataset of synthesized and ground truth patches and an appropriate loss function~\cite{johnson2016perceptual, zhao2017loss}.
However, at inference time, this approach would only leverage generic image priors and disregard the valuable information the input images carry.
Instead, we turn to the depth probability volume.
Consider the case of a ray traveling close to a depth discontinuity, which is likely to generate artifacts.
The probability distribution along this ray generally shows a peak corresponding to the foreground and one to the background, see Figure~\ref{fig:depth_prob}.
Then, rather than fixing the artifacts only based on generic image priors, we can guide the refinement network with patches extracted from the input views at the locations reprojected from these depths.
Away from depth discontinuities, the distribution usually has a single, strong peak, and the synthesized images are generally correct.
Still, since we warp the pixels independently, slight depth inaccuracy may cause local deformation.
Once again, patches from the input views can inform the refinement network about the underlying structure even if the depth is slightly off.

To minimize view-dependent differences in the patches without causing local deformations, we warp them with the homography induced by the depth plane.
For a given disparity $d=\bar{d}$, we compute the warped patch
\begin{equation}
    \widehat{P}_{i,j} = \text{W}(P_{i,j},H^{d=\bar{d}}_{i\rightarrow \text{NV}}),
\end{equation}
where $\text{W}(\cdot,H)$ is an operator that warps a patch based on homography $H$, and $H^{d=\bar{d}}_{i\rightarrow \text{NV}}$ is the homography induced by plane at disparity $\bar{d}$.
This patch selection strategy can be seen as an educated selection of a plane sweep volume~\cite{collins1996space}, where only the few patches that are useful are fed into the refinement network, while the large number of irrelevant patches, which can only confuse it, are disregarded.
In the next section we describe our refinement network, as well as details about its training.

\subsection{Refinement Network}\label{sec:refinement_network}
\begin{figure}[tb]
    \centering
    \includegraphics[width=.75\columnwidth,trim={0 15 0 0}, clip]{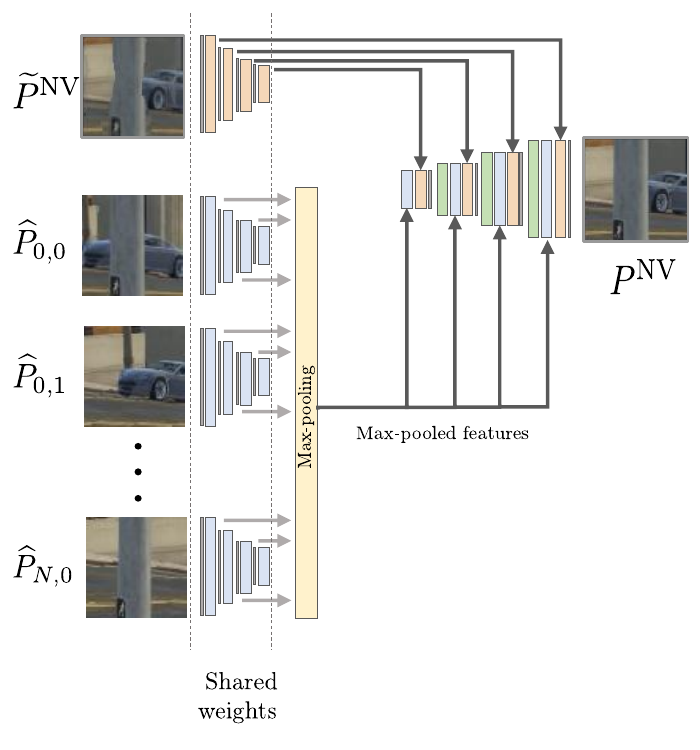}
    \caption{The refinement network takes as input a patch $\widetilde{P}^\text{NV}$ from the synthesized image $\NVI$, and a variable number of warped patches $\widehat{P}_{i,j}$ from each input view $\I_i$. All patches go through an encoder network. The features of the warped patches are aggregated using max-pooling. Both feature sets are concatenated and used in the decoder that synthesizes the refined patch $P^\text{NV}$.}
    \label{fig:refinement_patches}
    \vspace{-3mm}
\end{figure}

Our refinement strategy, shown in Figure~\ref{fig:refinement_patches}, takes a synthesized patch $\widetilde{P}^\text{NV}$ and $J$ warped patches $\widehat{P}_{i,j}$ from each input view $\I_i$.
The number of patches contributed to each $\widetilde{P}^\text{NV}$ can change from view to view: because of occlusions, an input image may not ``see'' a particular patch, or the patch could be outside of its field of view.
Moreover, the depth distribution along a ray traveling close to a depth discontinuity may have one peak, or several.
As a result, we need to design our refinement network to work with a variable number of patches.

\paragraph{Network Architecture.}
We use a UNet architecture for its proven performance on a large number of vision applications.
Rather than training it on a stack of concatenated patches, which would lock us into a specific value of $J$, we apply the encoder to each of the available patches independently. We then perform max-pooling over the features generated from all the available patches and we concatenate the result with the features of the synthesized patch, see Figure~\ref{fig:refinement_patches}.
The encoder has seven convolutional layers, four of which downsample the data by means of strided convolution.
We also use skip connections from the four downsampling layers of the encoder to the decoder.
Each skip connection is a concatenation of the features of the synthesized patch for that layer and a max-pooling operation on the features of the candidate patches at the same layer.

\paragraph{Training.}
We train the refinement network using the MVS-Synth dataset~\cite{huang2018deepmvs}.
We use a perceptual loss~\cite{johnson2016perceptual} as done by Zhuo~\etal~\cite{zhou2018stereo}, and train with ADAM~\cite{kingma2014adam}.
More details about the network and the training are in the Supplementary.

\section{Evaluation and Results}\label{sec:results}

\begin{figure*}[tb]
    \centering
    \includegraphics[width=\textwidth]{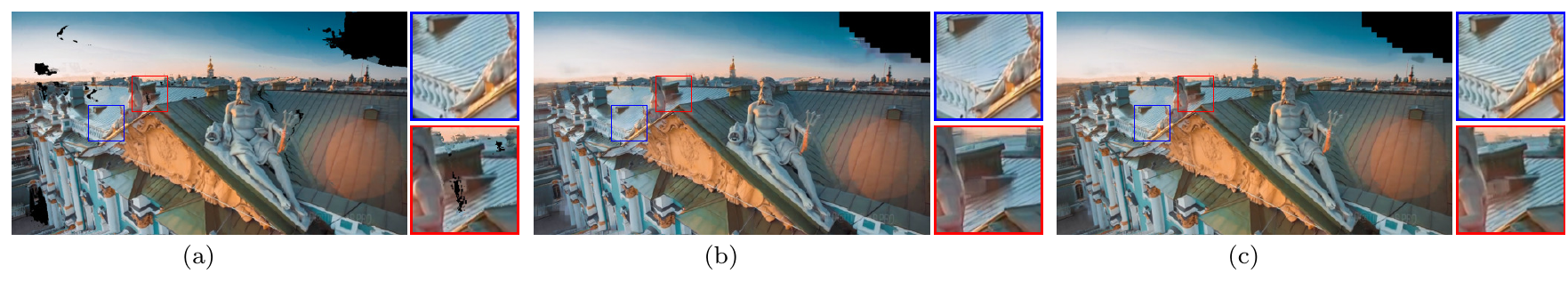}\
    \includegraphics[width=\textwidth]{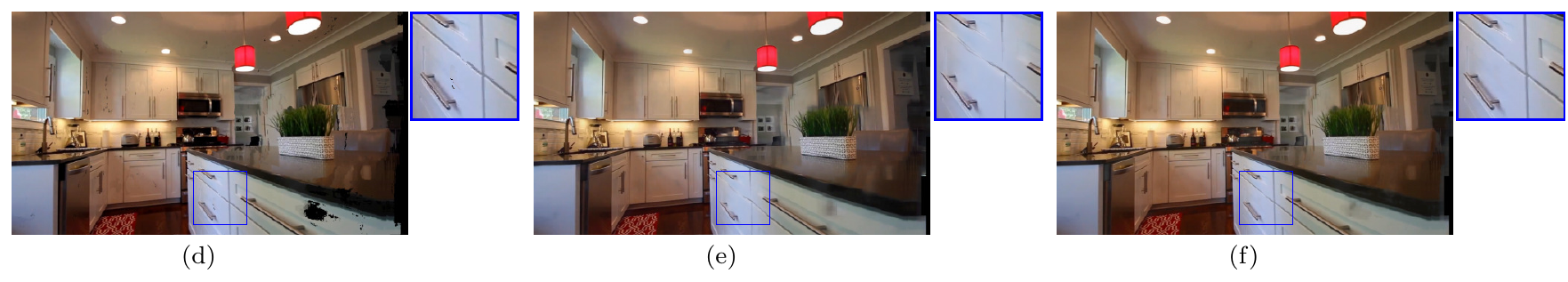}
    \caption{Our refinement network leverages information from relevant patches from the input images. Here (a) and (d) are $\widetilde{\mathcal{I}}^\text{NV}$, (b) and (e) are images created by training the network to refine patch only based on image priors, (c) and (f) are our results, which use the patches $\widehat{P}_{i,j}$. The structure of the roof is sharper in (c) compared to (b) and (a). The kitchen cabinet is correctly rendered in (f) compared to (d) and (e). }
    \label{fig:vnp}
    \vspace{-3mm}
\end{figure*}

\begin{figure}[tb]
    \centering
    \animategraphics[width=\columnwidth,timeline=figures/animated/animation.txt]{2}{figures/animated/red/a_}{0000}{0012}
    \caption{Animation showing the first three scenes in Figure~\ref{fig:results_table}. Requires a media-enabled viewer such as Adobe Reader. {\bf Click on the image to start the animation.}}
    \label{fig:animated}
    \vspace{-3mm}
\end{figure}

In this section we offer a numerical evaluation of our method and present several visual results.
We recommend to zoom into the images in the electronic version of the paper to better inspect them, and to use a media-enabled PDF viewer to play the animated figures.

\paragraph{Execution Time.} Using two views as an input, the depth probability volumes take $40$s, view synthesis (estimation of the depth volume in the novel view and rendering) takes 30s, and the refinement network takes $28$s (all averages).

\paragraph{Synthetic Scenes.} Non-blind image quality metrics such as SSIM~\cite{SSIM} and PSNR require ground truth images.
For a quantitative evaluation of our proposed method we use the MVS-Synth~\cite{huang2018deepmvs} dataset.
The MVS-Synth dataset provides a set of high-quality renderings obtained from the game GTA-V, broken up into a hundred sequences.
For each sequence, color images, depth images, and the camera parameters for each are provided.
The location of the cameras in each sequence is unstructured.
In our evaluation we select two adjacent cameras as the input views to our method and generate a number of nearby views also in the sequence.
We then compute the PSNR and SSIM metrics between the synthesized and ground-truth images.

In addition, we can use the same protocol to compare against Stereo Magnification (SM) by Zhou~\etal\cite{zhou2018stereo}. 
Although SM is tailored towards magnifying the baseline of a stereo pair, it can also render arbitrary views that are not in the baseline between the two cameras. 
We chose to quantitatively compare against SM because it also addresses the problem of generating extreme views, although in a more constrained setting.

Table~\ref{tab:numCompSM} shows PSNR and SSIM values for our method before and after refinement, and for SM.
The results show that the refinement network does indeed improve the quality of the final result.
In addition, the metrics measured on our method output are higher than those of SM.

\begin{table}
\centering
\small
\begin{tabular}{c|c|c|c}
\hline
Metric & Ours Warped & Ours Refined & SM \\
\hline
Mean SSIM & 0.851 & \textbf{0.877} & 0.842 \\
Mean PSNR & 24.6dB & \textbf{27.38dB} & 25.49dB \\
\hline
\end{tabular}\caption{Quantitative analysis of our proposed method and SM. ``Ours warped" refers to the images produced by backward warping before refinement,``Ours refined" refers to the images created by the refinement network, and ``SM" refers to the images created by the method of Zhou~\etal\cite{zhou2018stereo}}  \label{tab:numCompSM}
\vspace{-3mm}
\end{table}

\paragraph{Real Scenes.}
While sequences of real images cannot be used to evaluate our algorithm numerically, we can at least use them for visual comparisons of the results.

We perform a qualitative evaluation and compare against SM on their own data.
In their paper, Zhou~\etal show results when magnifying a stereo baseline by a factor of $4.5\times$.
While their results are impressive at that magnification, in this paper we push the envelop to \emph{extreme} and show results for {$30\times$} magnification of the input baseline.

\begin{figure}[tb]
    \centering
    \includegraphics[width=\columnwidth]{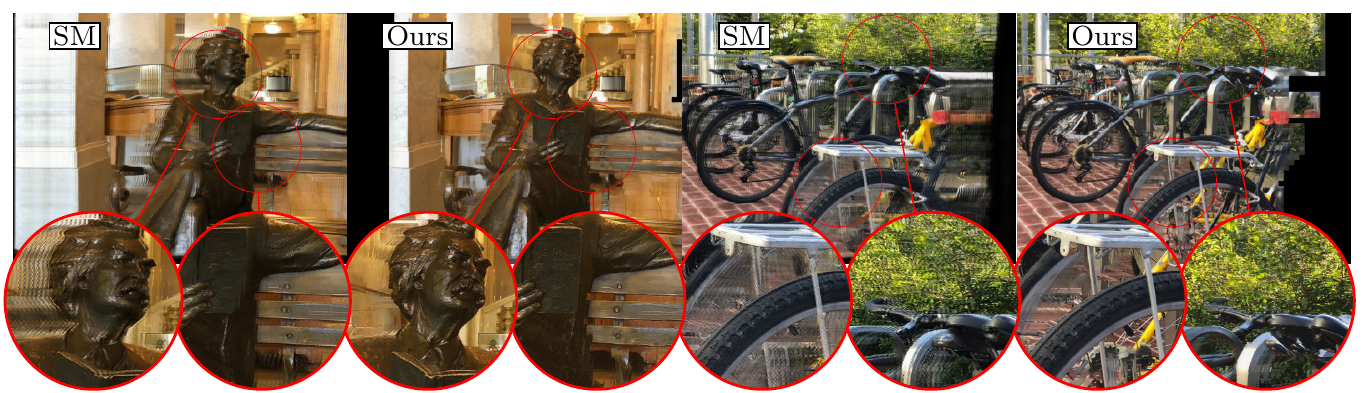}
    \caption{Comparison with Stereo Magnification for a $30\times$ baseline magnification. While some unavoidable artifacts are visible in both methods, our results have fewer, less noticeable artifacts, and are generally sharper. Please zoom in for the best viewing experience.}
    \label{fig:sm_comp}
    \vspace{-5mm}
\end{figure}

Figure~\ref{fig:teaser} and~\ref{fig:sm_comp} show {$30\times$} magnification on stereo pairs of scenes with complicated structure and occlusions.
At this magnification level, the results of Zhou~\etal are affected by strong artifacts. Even in the areas that appear to be correctly reconstructed, such as the head of Mark Twain's statue in Figure~\ref{fig:sm_comp}(left), a closer inspection reveals a significant amount of blur.
Our method generates results that are sharper and present fewer artifacts.
We also compare against their method at the magnification level they show, and observe similar results, see Supplementary.

\begin{figure}[tb]
    \centering
    \subfloat[]{\includegraphics[width=.48\columnwidth,trim={102 0 102 0},clip]{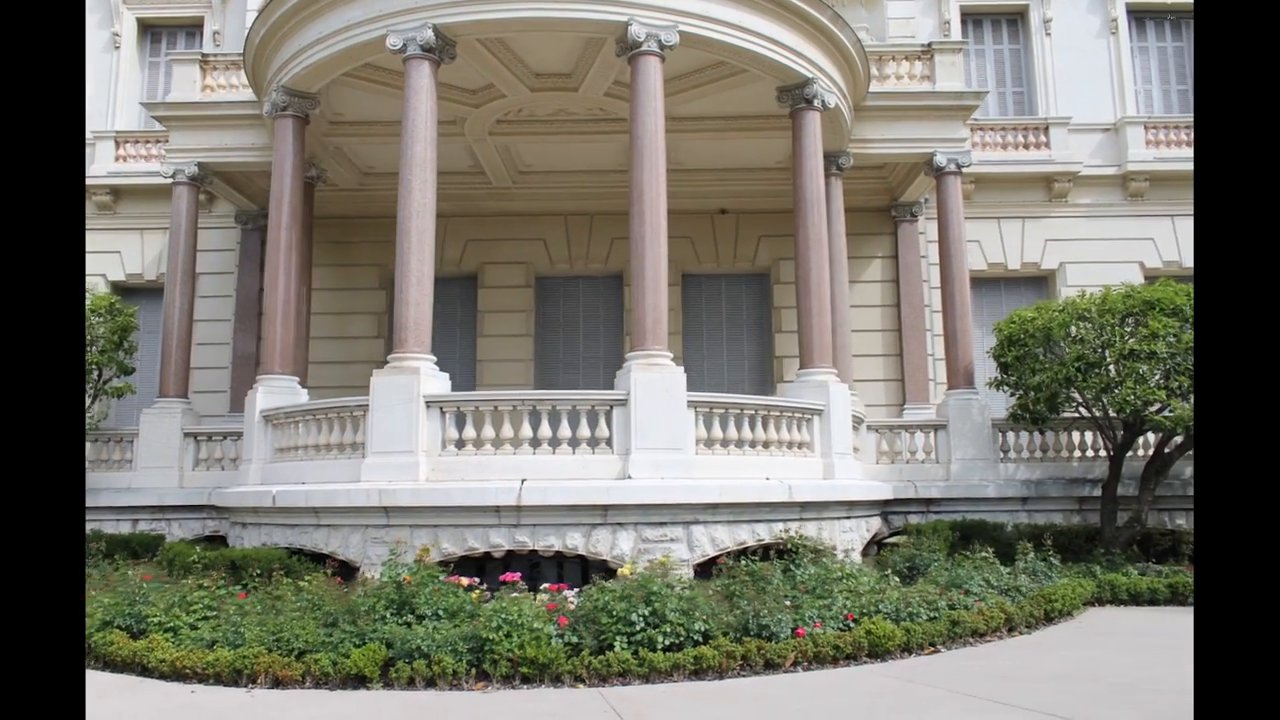}}~
    \subfloat[]{\includegraphics[width=.48\columnwidth]{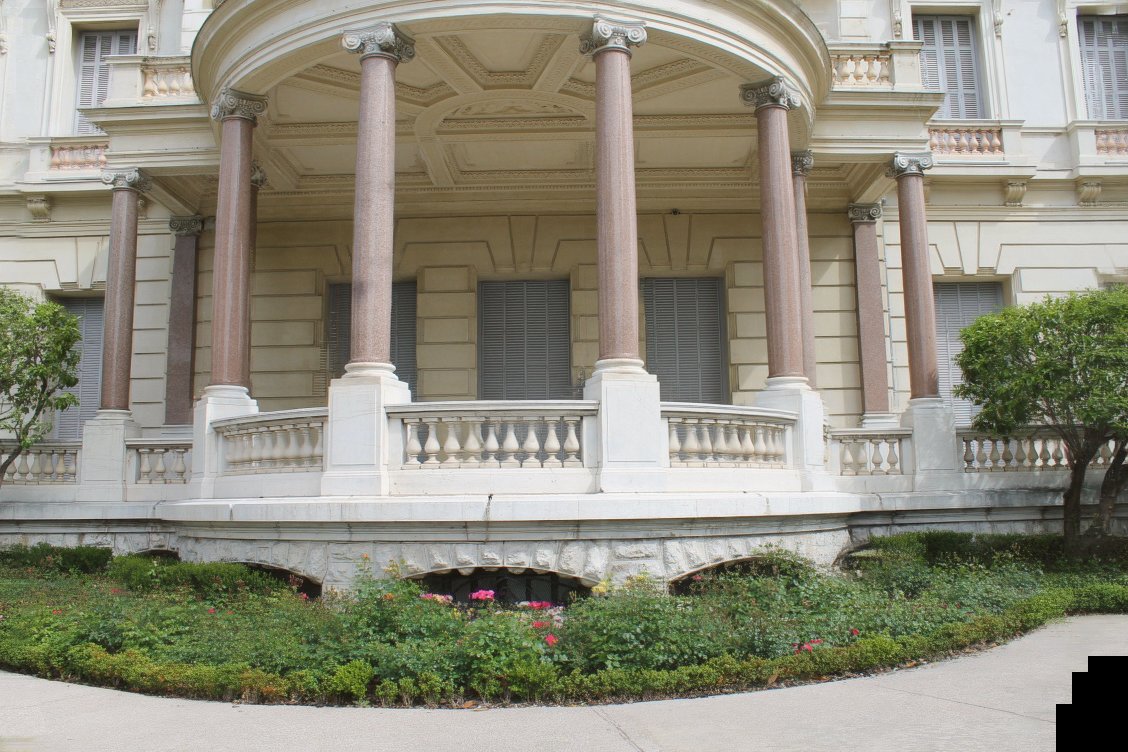}}
    \caption{Comparison with Soft3D~\cite{penner2017soft}. (a) A frame generated by Soft3D, which uses all the cameras in the sequence from Chaurasia~\etal~\cite{chaurasia2011silhouette}, and a frame generated by our method using only two input cameras around the middle of the sequence (b).}
    \label{fig:soft3D_comp}
    \vspace{-5mm}
\end{figure}

The method by Penner and Zhang arguably produces state-of-the-art results for novel view synthesis.
However, their code is not available and their problem setting is quite different in that they focus on interpolation and rely on a larger number of input cameras than our method.
For completeness, however, we show a comparison against their method in Figure~\ref{fig:soft3D_comp}.
Our reconstruction, despite using many fewer inputs, shows a quality that is comparable to theirs, though it degrades for larger extrapolation.

To validate our method more extensively, inspired by the collection strategy implemented by Zhou~\etal\cite{zhou2018stereo}, we capture a number of frame sequences from YouTube videos.

\begin{figure*}[tb]
	\centering
	\includegraphics[width=\textwidth]{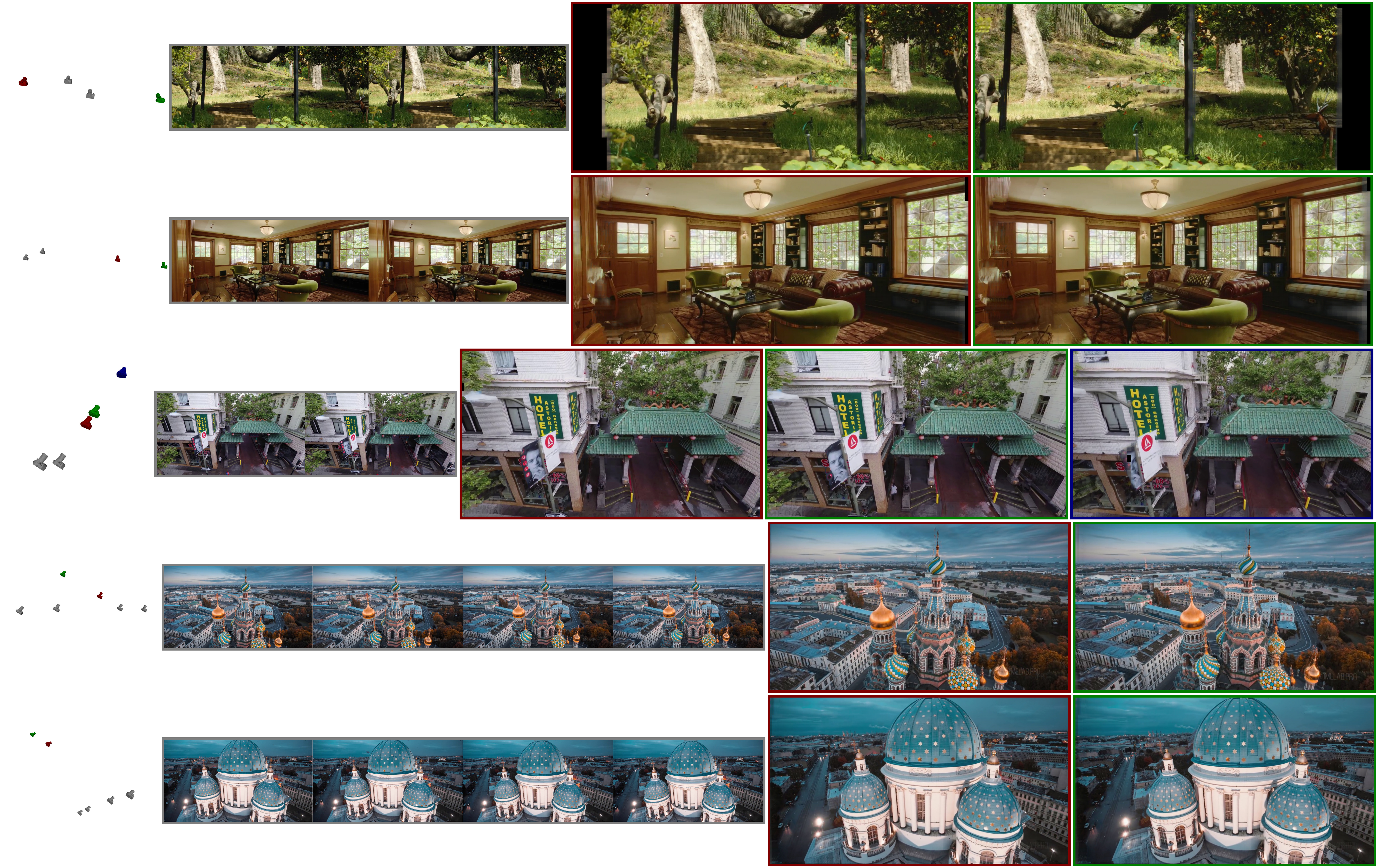}
	\caption{Extreme view synthesis on two-camera inputs and on four-camera inputs. For each row the cameras on the left show the position of the input views (light gray) and virtual views. The color of the pictures' frames matches the color of the corresponding camera on the left. The cameras on the left are rendered at the same scale to facilitate a comparison between the amounts of extrapolation in each case.}
	\label{fig:results_table}
    \vspace{-5mm}
\end{figure*}

A few of the results are shown in Figure~\ref{fig:results_table}. The leftmost column shows the camera locations for the images shown on the right.
The color of the cameras matches the color of the frame around the corresponding image, and gray indicates input cameras.
We present results for a number of different camera displacements and scenes, showcasing the strength of our solution.
In particular, the first three rows show results using only two cameras as inputs, with the virtual cameras being displaced by several times the baseline between the inputs cameras.
The third row shows a dolly-in trajectory (\ie, the camera moves towards the scene), which is a particularly difficult case.
Unfortunately, it may be challenging to appreciate the level of extrapolation when comparing images side by side, even when zooming in.
However, we also show an animated sequence in Figure~\ref{fig:animated}. To play the sequence, click on the image using a media-enabled reader, such as Adobe Reader. In the Supplementary we show additional video sequences and an animation that highlights the extent of parallax in one of the scenes.

Furthermore, our method can take any number of input images. The last two rows of Figure~\ref{fig:results_table} show two scenes for which we used four input cameras.

\paragraph{Refinement Network.}
We also conduct an evaluation to show that the use of patches as input to the refinement network does indeed guide the network to produce a better output.
Figure~\ref{fig:vnp} shows a comparison between our network and a network with the same exact number of parameters---the architecture differs only in the fact that it does not have additional patches.
It can be observed that the proposed architecture (Figure~\ref{fig:vnp}(c) and Figure~\ref{fig:vnp}(f)) can reconstruct local structure even when the single-patch network (Figure~\ref{fig:vnp}(b) and Figure~\ref{fig:vnp}(e)) cannot.
Indeed, the refinement network guided by patches can synthesize pixels in areas that had previously been occluded.

\subsection{Limitations}
While the refinement network can fix artifacts and fill in holes at the disocclusion boundaries, it can not hallucinate pixels in areas that were outside of the frusta of the input cameras---that is a different problem requiring a different solution, such as GAN-based synthesis~\cite{wang2018high}.
The refinement network  also struggles to fix artifacts that look natural, such as an entire region reconstructed in the wrong location.

Finally, because the depth values are discrete, certain novel views may be affected by depth quantization artifacts.
A straightforward solution is to increase the number of disparity levels (at the cost of a larger memory footprint and execution time) or adjust the range of disparities to better fit the specific scene.

\section{Conclusions}\label{sec:conclusions}

We presented a method to synthesize novel views from a set of input cameras.
We specifically target \emph{extreme} cases, which are characterized by two factors: small numbers of input cameras, as few as two, and large extrapolation, up to $30\times$ for stereo pairs.
To achieve this, we combine traditional geometric constraints with the learned priors.
We show results on several real scenes and camera motions, and for different numbers of input cameras.

\section*{Acknowledgments}
The authors would like to thank Abhishek Badki for his help with Figure~\ref{fig:results_table}, and the anonymous reviewers for their thoughtful feedback.

{\small
\bibliographystyle{ieee_fullname}
\bibliography{egbib}
}

\end{document}